\documentclass[11pt]{article}

\usepackage[preprint]{acl}
\usepackage{longtable}
\usepackage{times}
\usepackage{latexsym}

\usepackage[T1]{fontenc}

\usepackage[utf8]{inputenc}

\usepackage{microtype}

\usepackage{inconsolata}

\usepackage{graphicx}
\usepackage{amsmath,amssymb}
\usepackage{subcaption}
\usepackage{booktabs}
\title{Grounded Concreteness: Human-Like Concreteness Sensitivity in Vision–Language Models}

\author{
  Aryan Roy \\ Georgia Institute of Technology \\ \texttt{aroy389@gatech.edu}
  \And
  Zekun Wang \\ Georgia Institute of Technology \\ \texttt{zekun@gatech.edu}
  \AND
  Christopher J. MacLellan \\ Georgia Institute of Technology \\ \texttt{cmaclell@gatech.edu}
}

\begin{document}
\maketitle
\begin{abstract}
Do vision--language models (VLMs) develop more human-like sensitivity to linguistic concreteness than text-only large language models (LLMs) when both are evaluated with text-only prompts?
We study this question with a controlled comparison between matched Llama text backbones and their Llama Vision counterparts across multiple model scales, treating multimodal pretraining as an ablation on perceptual grounding rather than access to images at inference. 
We measure concreteness effects at three complementary levels: (i) output behavior, by relating question-level concreteness to QA accuracy; (ii) embedding geometry, by testing whether representations organize along a concreteness axis; and (iii) attention dynamics, by quantifying context reliance via attention-entropy measures. 
In addition, we elicit token-level concreteness ratings from models and evaluate alignment to human norm distributions, testing whether multimodal training yields more human-consistent judgments. 
Across benchmarks and scales, VLMs show larger gains on more concrete inputs, exhibit clearer concreteness-structured representations, produce ratings that better match human norms, and display systematically different attention patterns consistent with increased grounding.
\end{abstract}
\section{Introduction}
Human meaning is not uniformly ``linguistic'': some concepts are tightly linked to perception and action (e.g., \emph{apple, run}), while others are largely relational and context-dependent (e.g., \emph{stronger, justice}).  A long tradition in cognitive science treats \emph{concreteness} as a graded dimension of conceptual representation, with concrete words benefiting from richer sensory codes and exhibiting robust behavioral advantages over abstract words \citep{paivio1990mental,barsalou2008grounded}.  
Concreteness therefore offers a rare bridge between cognitive theory (how humans represent meaning) and computational diagnostics (how models encode and use meaning), enabling measurable tests of \emph{cognitive alignment} between humans and modern language systems \citep{Coltheart1981,brysbaert2014}. 
More broadly, recent work in language acquisition argues that neural models can serve as hypothesis generators and testers for cognitive theories, provided we design analyses that connect internal mechanisms to behavioral signatures \citep{portelance2022wordlearning}.

\begin{figure}[t]
    \centering
    \includegraphics[width=0.95\linewidth]{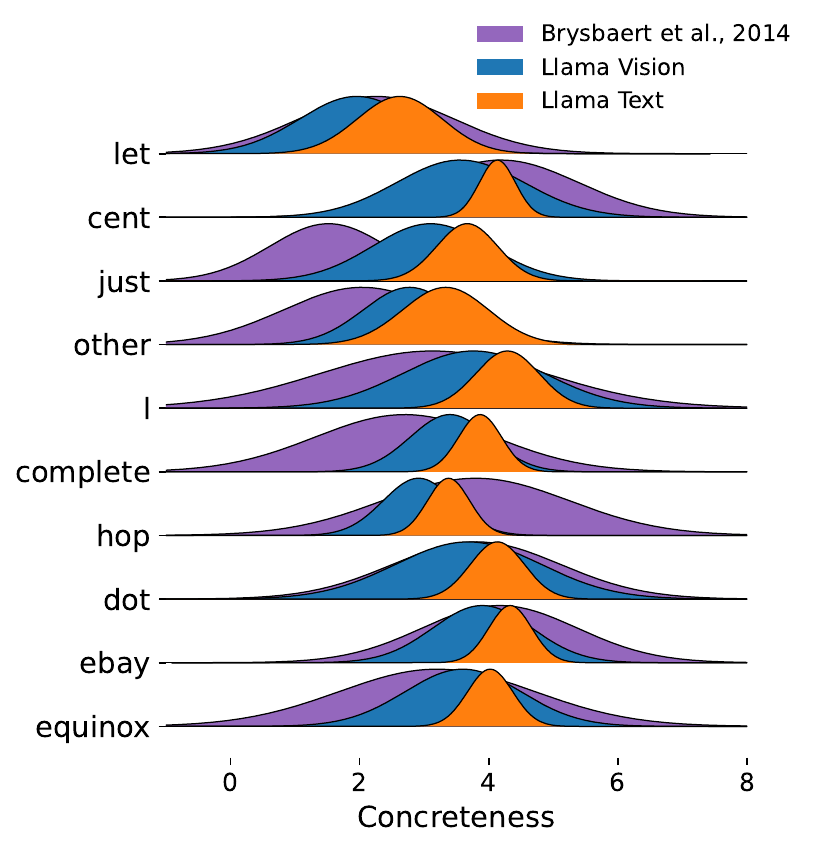}
    \caption{Comparison of concreteness rating distributions for selected words. For each word, we plot the empirical distribution of model-generated token ratings from Llama Vision (VLM) and Llama Text (LLM), alongside human norms from \citet{brysbaert2014}.}
    \label{fig:banner}
\end{figure}

A central tension is that contemporary large language models (LLMs) learn from text alone, raising questions about whether ``meaning'' can be recovered from form without grounding \citep{harnad1990symbol,bender-koller-2020-climbing}.  While distributional learning can capture many semantic regularities, the absence of perceptual experience may be especially consequential for \emph{concreteness}: in humans, concrete concepts are supported by sensorimotor simulations and imagery-like codes \citep{paivio1990mental,barsalou2008grounded}.  Vision-language models (VLMs) offer a natural testbed for this debate.  By aligning text with visual representations (e.g., CLIP-style contrastive learning and instruction-tuned VLMs), VLMs may develop more human-like, graded concreteness representations than comparably sized text-only LLMs \citep{radford2021learning,alayrac2022flamingo,liu2023visual,touvron2023llama}.  
Yet prior evidence connecting concreteness to model behavior and representations is difficult to interpret as a \emph{vision effect}.
On one hand, multimodal/distributional semantics work links concreteness to perceptual grounding and visual consistency \citep{hill-etal-2014-multi,hessel-etal-2018-quantifying,mickus-etal-2023-grounded};
on the other hand, separate lines probe concreteness as an axis in embedding spaces using text-derived features \citep{charbonnier-wartena-2019-predicting,wartena-2024-estimating}.
However, these strands rarely provide a controlled ablation that isolates the contribution of visual input, and they typically analyze a single level (task performance \emph{or} representations) rather than jointly linking behavior, geometry, and processing dynamics.

This motivates a controlled LLM--VLM ablation in which the language backbone is held as comparable as possible and the primary difference is access to vision, allowing the contribution of visual grounding to be isolated.
In this work, concreteness awareness is evaluated under this ablation by triangulating evidence across three complementary levels of analysis.
First, at the \textbf{output level}, we ask whether VLM accuracy on question answering increases with the concreteness of the queried concepts.
We further elicit model-produced concreteness ratings (token-level) and measure their alignment with human norms (Figure \ref{fig:banner}) \citep{Coltheart1981,brysbaert2014}.
Second, at the \textbf{embedding level}, we test whether token representations organize by graded concreteness by projecting token-level embeddings into a low-dimensional space (t-SNE) and measuring within-bin compactness via intra-cluster dispersion across concreteness bins.
Third, at the \textbf{attention level}, we quantify contextual dependence as the entropy of each token's self-attention distribution: if abstract meaning is more compositionally supported by surrounding context, abstract tokens should exhibit broader, higher-entropy attention, whereas concrete tokens should exhibit sharper, lower-entropy attention concentrated on fewer positions. 
This prediction aligns with classic context availability accounts of concreteness effects in psycholinguistics, which argue that abstract words benefit more from supportive context for comprehension than concrete words \citep{schwanenflugel-shoben-1983,schwanenflugel-akin-luh-1992}.



Our analyses are designed as a vision ablation for semantic grounding: holding the base language model family and scaling regime as constant as possible, we ask what changes when vision is introduced.  
This yields a coherent story linking classic cognitive accounts of concreteness---dual coding and grounded cognition \citep{paivio1990mental,barsalou2008grounded}---to measurable signatures in modern foundation models: (i) \emph{behavioral sensitivity} to concreteness in downstream QA, (ii) \emph{representational geometry} that recovers a concreteness ordering, and (iii) \emph{contextual dependence} reflected by attention entropy.


\paragraph{Contributions.}
We make following contributions:
(1) a controlled LLM--VLM ablation study that isolates the effect of visual grounding on concreteness awareness across matched model families and scales;
(2) output-level evidence that tests concreteness sensitivity in QA and quantifies alignment between model-elicited concreteness ratings and human norms;
(3) internal diagnostics connecting grounding-based accounts to model representations and processing, including concreteness-conditioned clustering in embedding space (t-SNE with intra-cluster dispersion) and attention-entropy measures of contextual dependence motivated by context-availability theories of abstract meaning.

\section{Related work}




\subsection{Measuring concreteness in language}
Concreteness is a graded psycholinguistic property that captures how directly a concept can be experienced through the senses, and it has been extensively measured through human norming studies.
Classic resources such as the MRC Psycholinguistic Database provide lexical attributes including concreteness/tangibility judgments \citep{Coltheart1981}, and later large-scale norms substantially expand coverage and improve reliability for modern evaluation settings \citep{brysbaert2014}.

In parallel, computational work has proposed algorithmic approximations of concreteness.
Early NLP approaches connected concreteness-related cues to figurative language phenomena, using concrete vs.\ abstract contextual signals for literal/metaphorical sense identification \citep{turney-etal-2011-literal}.
More recent methods treat concreteness prediction as a supervised estimation problem over distributional features and contextual representations \citep{charbonnier-wartena-2019-predicting}.
A particularly relevant line incorporates visual information: ``visual concreteness'' can be operationalized via cross-image consistency within multimodal datasets \citep{hessel-etal-2018-quantifying}, and visually grounded learning objectives can shape linguistic structure and representations \citep{shi-etal-2019-visually}.
Given known context sensitivity (a word’s perceived concreteness can shift with discourse and reference), human norms provide an external anchor for evaluation, while model-produced ratings can be treated as context-conditioned distributions rather than fixed type-level attributes.

\subsection{Grounding and vision language models}
Grounding-based accounts of meaning emphasize that linguistic symbols ultimately connect to perception and action, motivating the classic symbol-grounding problem \citep{harnad1990symbol}.
In NLP, grounding has been studied through both definitional discussions and benchmark/task design, with critiques highlighting that ``grounding'' can mean different things depending on modality, interaction, and evaluation protocol.
This motivates evaluating grounding beyond downstream success rates, using complementary diagnostics that probe internal representations and processing rather than relying on task behavior alone \citep{bisk-etal-2020-experience,chandu-etal-2021-grounding,mickus-etal-2023-grounded}.



Modern VLMs provide a scalable route to grounding by learning joint representations of text and vision.
Several models are explicitly designed to encourage fine-grained grounding via cross-modal alignment objectives (e.g., word/phrase--region alignment) \citep{tan-bansal-2019-lxmert,chen-etal-2020-uniter}, and to evaluate grounded lexical acquisition beyond standard downstream transfer \citep{ma-etal-2023-world}.
Large-scale contrastive and generative pretraining has produced general-purpose models that align linguistic descriptions with visual features, supporting transfer to many multimodal tasks \citep{radford2021learning,alayrac2022flamingo,liu2023visual}.

These strands motivate treating vision as a causal factor that can strengthen concreteness awareness.
If concrete concepts are more consistently tied to perceptual regularities (i.e., they have more stable visual correlates), then adding visual supervision should preferentially benefit how models recognize and represent concreteness.

\subsection{Neural models as cognitive probes of language learning and processing}
A growing cognitive-science perspective treats neural networks as tools for generating and testing mechanistic hypotheses about human learning, rather than purely as engineering solutions \citep{Portelance2022}.
This includes using language models as ``psycholinguistic subjects,'' evaluating whether model-based surprisal and state representations predict human processing difficulty and syntactic expectations \citep{goodkind-bicknell-2018-predictive,futrell-etal-2019-neural}.
Another line tests whether models acquire human-relevant grammatical generalizations via targeted syntactic evaluations and minimal-pair benchmarks \citep{linzen-etal-2016-assessing,gulordava-etal-2018-colorless,warstadt-etal-2020-blimp-benchmark}.
More recent work pushes toward developmental plausibility by constraining data and supervision (e.g., BabyLM) or by studying interactive learning dynamics \citep{warstadt-etal-2023-babylm,ma-etal-2025-babysit}.
Together, these efforts motivate treating representational properties of language models as empirical objects for studying human cognitive constructs.

\section{Experiment Setup}

\paragraph{Models}
The study compares matched pairs of text-only LLMs and vision-language models (VLMs) at two parameter scales. For the LLMs, the backbone is Meta's Llama 3.1 family (an 8B and a 70B Text-only Model) \citep{meta2024llama31}. For VLMs, the corresponding vision models from the Llama Vision 3.2 family (an 11B and a 90B vision LLM) \cite{meta2024llama32} were chosen that utilize the Llama 3.1 text-only models as a backbone. The text-only models are fitted with a vision adapter and then trained on a multimodal dataset to create the vison models. 
We refer Appendix \ref{app:model_details} for details. 
Unless otherwise stated, evaluation uses text-only prompts (no images), so the LLM--VLM comparison functions as an ablation on \emph{access to visual supervision during training} rather than access to images at inference.

\paragraph{Measuring concreteness}
Token-level concreteness $C(w)$ is obtained from the human ratings in the 40k English words \cite{brysbaert2014} (40K).
For each word that appears in 40K, its concreteness score is set to the corresponding 40K mean rating. 
For out-of-vocabulary proper nouns (e.g., named entities) that are not covered by 40K, the score is set to the maximum concreteness value on the 40K scale (5). 
For function words without a clear concreteness interpretation (e.g., articles and prepositions) that are also absent from 40K, the score is set to $0$. 
Sentence-level concreteness for an input string $x$ with word tokens $w_{1:n}$ is the mean of token scores:
\begin{equation}
C(x)=\frac{1}{n}\sum_{i=1}^{n} c(w_i).
\end{equation}
For subword-tokenized model inputs, word-level scores are propagated to constituent sub-tokens to enable tokenwise analyses.

\paragraph{Text datasets}
Evaluation uses standard text-only QA benchmarks covering diverse reasoning demands and a broader range of question concreteness: ARC-Easy and ARC-Challenge for grade-school science multiple-choice questions \citep{clark2018arc}; BoolQ for naturally occurring yes/no questions \citep{clark-etal-2019-boolq}; WinoGrande for adversarial pronoun/coreference resolution \citep{sakaguchi2020winogrande}; CommonsenseQA for commonsense multiple-choice QA \citep{talmor-etal-2019-commonsenseqa}; Social IQA for reasoning about social interactions and implications \citep{sap-etal-2019-social} and PIQA for physical commonsense reasoning \citep{bisk2020piqa}. 
Performance is measured by accuracy under a unified prompting format.
We refer Appendix \ref{app:datasets} and \ref{app:prompts} for details on datasets and prompts.

\subsection{Research questions and methods}





This section describes how each hypothesis is operationalized and tested. All analyses are conducted for both model scales to assess scaling effects. 

\paragraph{Does the VLM outperform the LLM on QA questions as question concreteness increases?}
For each benchmark dataset, each question is scored as correct or incorrect under a unified prompting format.
To summarize performance as a function of concreteness, sentence-level concreteness scores are pooled across all datasets and discretized into six equal-width bins of size $0.6$, spanning $[1.8,\,4.8]$ (the observed range is $[1.96,\,4.67]$).
For each bin, accuracy is computed as the mean correctness over questions whose sentence concreteness falls in that interval.
In addition, the bin-wise accuracy gap between the VLM and its matched LLM is reported, $\Delta\mathrm{Acc}=\mathrm{Acc}_{\mathrm{VLM}}-\mathrm{Acc}_{\mathrm{LLM}}$, to quantify where vision provides an advantage.
We hypothesize that $\Delta\mathrm{Acc}$ is expected to be larger in higher-concreteness bins, indicating that the VLM is relatively more robust on concrete questions than the text-only model.


\paragraph{Do VLM token representations exhibit tighter within-concreteness clusters than LLM token representations?}
Each word ($w$) covered by 40K is rounded to a discrete concreteness bin $b(w)\in\{1,\dots,5\}$.
For each model, we extract last-layer contextual representations and average over occurrences to obtain a type vector $\bar{\mathbf{h}}(w)$.
To measure within-bin dispersion, we fit $\bar{\mathbf{h}}(w)$ with 2D t-SNE, yielding $\mathbf{z}(w)$. 
Within this t-SNE space, dispersion is measured as the mean pairwise cosine distance among tokens with the same label, where lower values indicate more compact clusters:
\begin{equation}
\label{eq:intra}
D \;=\; \mathbb{E}_{\ell}\,\mathbb{E}_{w\neq w'\sim\mathcal{W}_\ell}\!\left[1-\cos\!\bigl(\mathbf{z}(w),\mathbf{z}(w')\bigr)\right]
\end{equation}
where $\mathcal{W}_\ell$ represents the discretize concreteness bin. Lower $D$ indicates tighter within-concreteness clusters.

\paragraph{Do abstract tokens exhibit higher-entropy attention distributions than concrete tokens, and is this abstract--concrete separation sharper in VLMs?}
For each layer $\ell$ and head $h$, self-attention weights follow the standard Transformer definition \citep{vaswani2017attention}:
\begin{equation}
\mathbf{A}^{(\ell,h)} = \mathrm{softmax}\!\left(\frac{\mathbf{Q}^{(\ell,h)}\mathbf{K}^{(\ell,h)\top}}{\sqrt{d_k}}\right),\nonumber
\end{equation}
where $\mathbf{A}^{(\ell,h)}_{i,j}$ is the attention paid by token $i$ to token $j$ and forms a probability distribution over $j$ due to softmax normalization.
For each token $i$, attention entropy is computed as:
\begin{equation}
\label{eq:h}
H^{(\ell,h)}(i) = -\sum_j \mathbf{A}^{(\ell,h)}_{i,j}\log \mathbf{A}^{(\ell,h)}_{i,j}.
\end{equation}
Entropy is then averaged across heads for each layer to obtain a per-token entropy score.
At each layer, we test the association between token concreteness $c(w_i)$ and attention entropy via Pearson's $r$.
We expect a \emph{negative} correlation ($r<0$): abstract tokens should exhibit \emph{higher} attention entropy (more diffuse context integration), whereas concrete tokens should exhibit \emph{lower} entropy (more focused attention), consistent with concreteness effects in comprehension \citep{schwanenflugel-shoben-1983,schwanenflugel-akin-luh-1992}.
Moreover, we predict this effect is stronger in VLMs than LLMs (i.e., more negative $r$ in VLMs), reflecting a sharper abstract--concrete separation.

\paragraph{Do model generated concreteness judgments align better with human norms for VLMs?}
Each model is prompted to output a concreteness rating for every word token in each question on the 40K scale.
We refer Appendix \ref{app:prompts} for prompt details.
To elicit reliable concreteness judgments and enforce a consistent output format, we use the \texttt{instruct} variants of the larger models (70B text-only and 90B vision-language).
Because the same word type can appear in multiple contexts, each word $w$ induces an empirical distribution over ratings under a model $m$:
\begin{equation}
p_m(r\mid w)\propto \sum_{x\in\mathcal{X}(w)} \mathbf{1}\!\left[\hat r_m(w,x)=r\right]
\end{equation}
where $\mathcal{X}(w)$ are contexts containing $w$ and $\hat r_m(w,x)$ is the model-produced rating for token $w$ in question $x$.
We construct an analogous human distribution $p_H(r\mid w)$ from the 40K annotations and quantify human--model agreement with the symmetric KL divergence:
\begin{equation}
\label{eq:dkl}
D_{\mathrm{KL}}(w)=\tfrac12\!\left[\mathrm{KL}(p_m\|p_H)+\mathrm{KL}(p_H\|p_m)\right]
\end{equation}
where smaller $D_{\mathrm{KL}}(w)$ indicates better alignment.
We expect (i) $D_{\mathrm{KL}}(w)$ decreases as human concreteness increases, and (ii) this decrease is steeper for VLMs than for LLMs.
To test the trend, we bin words by human concreteness in 0.5-wide bins and regress binned $D_{\mathrm{SKL}}$ on bin center, reporting slope, $R^2$, and $p$-value.

\section{Analysis and discussion}

\begin{figure}[t]
    \centering
    \begin{subfigure}[t]{0.5\linewidth}
        \centering
        \includegraphics[width=\linewidth]{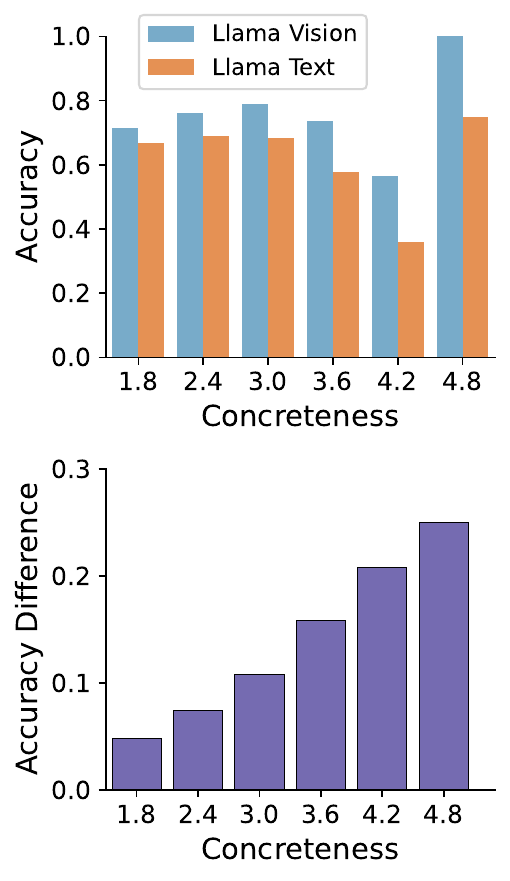}
        \caption{8B model family.}
        \label{fig:accuracy}
    \end{subfigure}\hfill
    \begin{subfigure}[t]{0.5\linewidth}
        \centering
        \includegraphics[width=\linewidth]{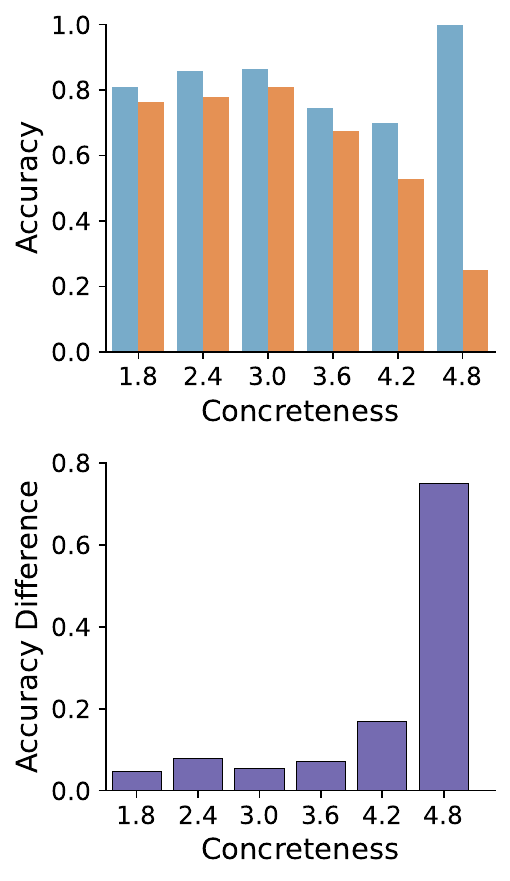}
        \caption{70B model family.}
        \label{fig:accuracy-large}
    \end{subfigure}
    \caption{Top row: accuracy by question concreteness for Llama Text vs.\ Llama Vision. Bottom row: the VLM--LLM accuracy gap.}
    \label{fig:accuracy-side-by-side}
\end{figure}

\paragraph{VLM outperform their LLM counterpart on QA questions and the gaps widen on more concrete questions.}
Figure~\ref{fig:accuracy-side-by-side} shows that, across all concreteness bins, VLMs consistently outperform their text-only counterparts at both scales.
Across all datasets, for the smaller pair, accuracy increases from 68.0\% (LLM) to 77.5\% (VLM), and for the larger pair from 78.8\% (LLM) to 85.5\% (VLM). 
Since each VLM is trained from the same model family as its LLM counterpart, these gains indicate that multimodal training transfers to improved textual QA.
We refer Appendix \ref{app:results} for additional per-dataset results.

Crucially, the advantage is not uniform across question types: the bottom row of Figure~\ref{fig:accuracy-side-by-side} shows that the VLM--LLM gap increases with question concreteness in both scales, with the strongest separation in the most concrete bin. 
This pattern suggests that visual grounding disproportionately benefits questions whose successful resolution depends on perceptible entities, attributes, and events (e.g., shape, material, spatial relations), consistent with grounded accounts in which perceptual experience provides an additional scaffold for semantic representations \citep{harnad1990symbol,barsalou2008grounded,bisk-etal-2020-experience}. 
A plausible mechanism is that vision--language training strengthens the association between concrete lexical items and perceptually anchored image features (e.g., object properties and spatial configurations), making the relevant evidence easier to retrieve and compose when answering concrete questions.
In other words, the VLM’s gains appear concentrated where the QA signal can be supported by grounded semantics rather than purely symbolic co-occurrence.

At the same time, the effect is smaller (and sometimes flatter) in lower-concreteness bins, where performance may depend more on abstract relations, discourse-level inference, or world knowledge not directly supported by perceptual grounding.
Moreover, abstract language tends to be more polysemous and context-dependent, which can reduce the benefit of any single additional modality.
Together, these results suggest that multimodal training provides an asymmetric benefit: it reliably improves QA overall, but disproportionately improves the processing and use of concrete concepts, which we further probe via representation geometry and attention diagnostics in subsequent experiments.

\paragraph{VLM token representations form tighter within-concreteness clusters than LLMs.}

\begin{figure*}[!th]
    \centering
    \begin{subfigure}[t]{1.0\linewidth}
        \centering
        \includegraphics[width=\linewidth]{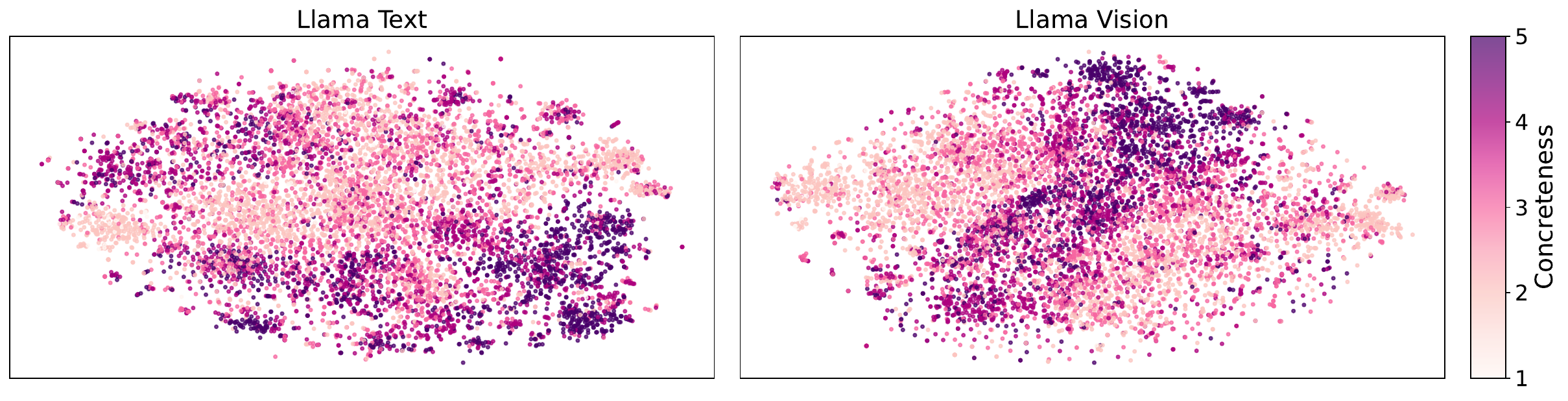}
        \caption{8B model family.}
        \label{fig:token-tsne}
    \end{subfigure}
    \begin{subfigure}[t]{1.0\linewidth}
        \centering
        \includegraphics[width=\linewidth]{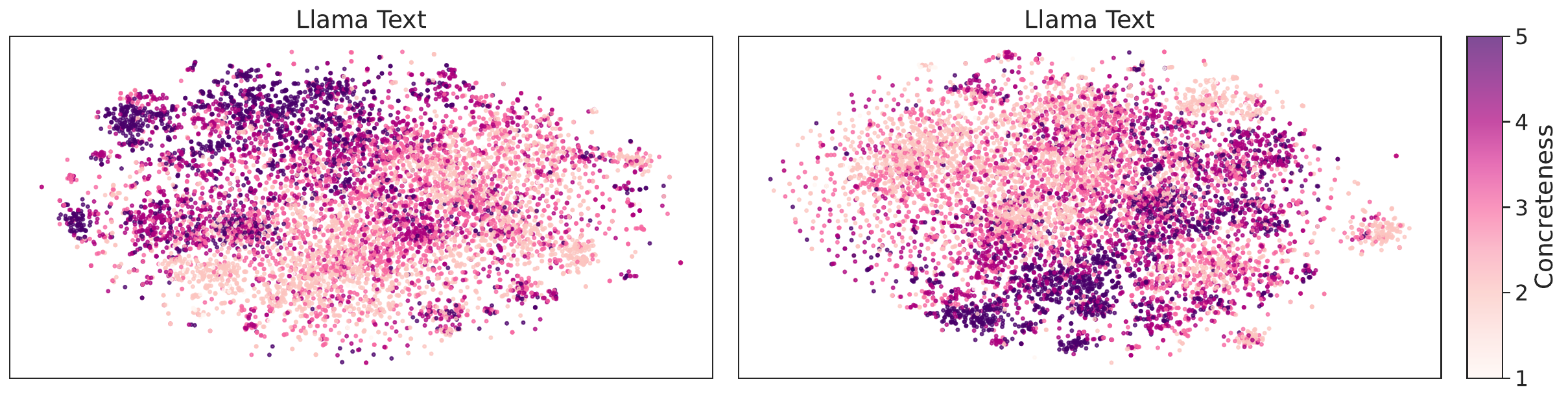}
        \caption{70B model family.}
        \label{fig:token-tsne-large}
    \end{subfigure}
    \caption{t-SNE of average last-layer token representations for Llama Text vs.\ Llama Vision, colored by human concreteness.}
    \label{fig:token-tsne-vertical}
\end{figure*}

Given that VLMs outperform their text-only counterparts on QA, we ask whether vision-text training also reshapes the \emph{geometry} of token representations along a graded concreteness dimension.
Figure~\ref{fig:token-tsne} provides a qualitative view: in both the smaller and larger models, the VLM embedding shows a visibly more contiguous band of highly concrete tokens (darker points), whereas the text-only LLM distributes high-concreteness tokens more diffusely across the map.
This pattern suggests that visual grounding encourages representations of perceptually grounded words to occupy a more coherent subregion of the space, consistent with grounded accounts of meaning and the symbol-grounding perspective \citep{harnad1990symbol,barsalou2008grounded,bisk-etal-2020-experience}.

We quantify this effect using within-bin intra-cluster dispersion (Eq.~\ref{eq:intra}) computed in the t-SNE space.
Table~\ref{tab:intra-dispersion} shows that VLMs achieve lower dispersion than LLMs at \emph{every} concreteness level in both families. The effect is largest for the most concrete bin ($c{=}5.0$): dispersion drops from 0.76$\rightarrow$0.66 in the 8B family and from 0.87$\rightarrow$0.77 in the 70B family.
Notably, dispersion is highest in the mid-concreteness range (roughly $c\in[2,4]$) and drops sharply for the most concrete words, which is compatible with the idea that mid-range words are more heterogeneous (e.g., broader senses or mixed perceptual/abstract usage) while highly concrete words admit more stable, visually grounded semantics.

Taken together, the qualitative structure in Figure~\ref{fig:token-tsne} and the consistent quantitative reductions in Table~\ref{tab:intra-dispersion} support our hypothesis that VLM representations encode graded concreteness more cleanly than LLMs.
This geometry offers a representational account of the concreteness-dependent QA gains. 
If highly concrete word types occupy a tighter region of the space, their representations are more consistent across contexts, reducing the need for context-dependent disambiguation and making grounded attributes easier to retrieve and compose. 
In turn, this should improve robustness on questions that hinge on perceptual properties (e.g., materials, shapes, spatial relations).


\begin{table}[t]
\centering
\small
\setlength{\tabcolsep}{5pt}
\begin{tabular}{@{}c cc cc@{}}
\toprule
 & \multicolumn{2}{c}{\textbf{8B model family}} & \multicolumn{2}{c}{\textbf{70B model family}} \\
\cmidrule(lr){2-3}\cmidrule(lr){4-5}
\textbf{Conc.} & \textbf{Text-only} $\downarrow$ & \textbf{Vision} $\downarrow$ & \textbf{Text-only} $\downarrow$ & \textbf{Vision} $\downarrow$ \\
\midrule
1.0 & 0.87 & \textbf{0.75} & 0.93 & \textbf{0.82} \\
2.0 & 0.94 & \textbf{0.87} & 0.98 & \textbf{0.94} \\
3.0 & 0.98 & \textbf{0.96} & 1.00 & \textbf{0.99} \\
4.0 & 0.99 & \textbf{0.96} & 1.00 & \textbf{0.99} \\
5.0 & 0.76 & \textbf{0.66} & 0.87 & \textbf{0.77} \\
\bottomrule
\end{tabular}
\caption{Within-concreteness cluster dispersion (mean pairwise cosine distance in 2D t-SNE; lower is tighter).}
\label{tab:intra-dispersion}
\end{table}

\paragraph{Concrete tokens show lower attention entropy, with a stronger effect in VLMs.}

\begin{figure}[!th]
    \centering
    \begin{subfigure}[t]{1.0\linewidth}
        \centering
        \includegraphics[width=\linewidth]{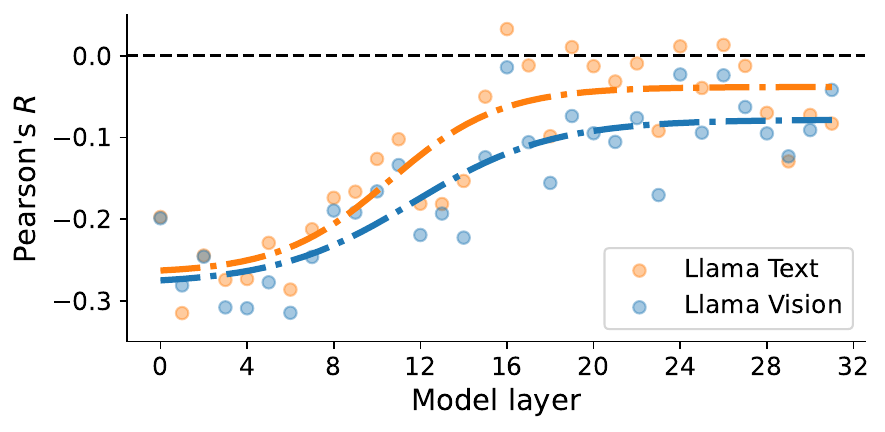}
        \caption{8B model family.}
        \label{fig:attention-r}
    \end{subfigure}
    \begin{subfigure}[t]{1.0\linewidth}
        \centering
        \includegraphics[width=\linewidth]{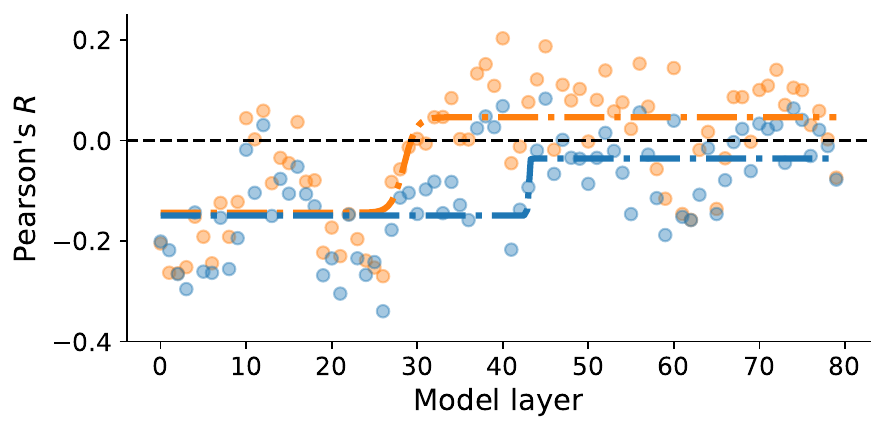}
        \caption{70B model family.}
        \label{fig:attention-r-large}
    \end{subfigure}
    \caption{Layerwise Pearson's $r$ between token concreteness and head-averaged attention entropy. Colored dash lines are sigmoid fitted curves.}
    \label{fig:attention}
\end{figure}

Motivated by the representational results showing tighter within-concreteness clustering in VLMs, we next test whether models also differ in how much context they integrate for abstract versus concrete words. 
Prior psycholinguistic work suggests that concrete words tend to be easier to interpret and rely less on contextual support than abstract words, which are more context-dependent \citep{schwanenflugel-shoben-1983,schwanenflugel-akin-luh-1992}. 
We operationalize contextual reliance using attention entropy (Eq.~\ref{eq:h}): higher entropy corresponds to more diffuse attention over many tokens, while lower entropy reflects more concentrated attention.

Figure~\ref{fig:attention} plots, for each layer, Pearson's $r$ between token concreteness and token attention entropy (head-averaged), for both model scales. 
We report full Pearson's $R$ values and their $p-$values in Appendix \ref{sec:addi-entropy}.
Across most layers, correlations are negative (or near zero), indicating that more concrete tokens exhibit lower attention entropy (i.e., they attend more selectively) whereas abstract tokens show higher entropy, consistent with the hypothesis that abstract meaning requires broader contextual integration. 
The effect is strongest in earlier-to-mid layers: both model families display more negative correlations in roughly the first half of the network, followed by a gradual attenuation toward later layers. 
This layerwise result suggests that context concreteness sensitivity is primarily expressed early in processing, while later layers may shift toward task-level integration that is less directly tied to lexical concreteness.

Importantly, VLMs show a consistently stronger negative correlation than their text-only counterparts. 
Averaging Pearson's $r$ across layers yields $r=-0.12$ (LLM) vs.\ $r=-0.16$ (VLM) for the smaller models, and $r=-0.02$ (LLM) vs.\ $r=-0.10$ (VLM) for the larger models, indicating a sharper abstract--concrete separation in VLM attention behavior,
One interpretation is that vision-text training provides an additional grounding signal that stabilizes the representations of concrete words, allowing the model to resolve them with more focused attention (lower entropy) and reducing the need to distribute attention broadly across other tokens.

\paragraph{VLM concreteness ratings align more closely with humans than LLM as concreteness increases.}

\begin{figure*}
    \centering
    \includegraphics[width=1\linewidth]{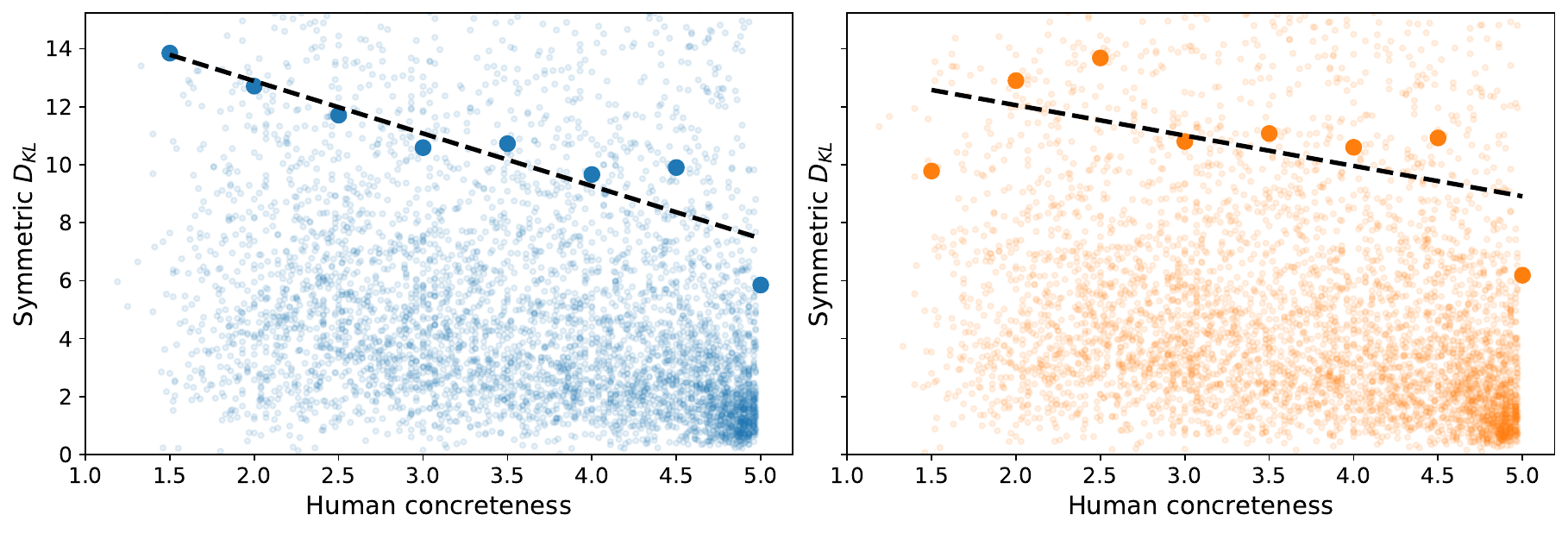}
    \caption{Human--model alignment of token-level concreteness judgments. Each point is a word with human concreteness on the x-axis and symmetric KL divergence between the model’s rating distribution and the human (40K) distribution on the y-axis (lower is better). Larger dots show bin averages; dashed lines are linear fits over bins. The VLM (left) exhibits lower divergence and a steeper decrease in divergence with concreteness than the matched text-only LLM (right).}
    \label{fig:rating}
\end{figure*}


Finally, if the performance, representation geometry, and attention analyses reflect a human-like graded concreteness sensitivity, we should also observe human-aligned concreteness judgments from the models. 
Figure~\ref{fig:rating} plots human concreteness against human--model agreement measured by symmetric KL divergence (lower is better). 
Overall, the VLM exhibits lower divergence than the LLM (mean $D_{\mathrm{KL}}$: 9.4 vs.\ 10.1), indicating closer alignment to human norms from \cite{brysbaert2014}.

More importantly, agreement improves with concreteness: as human ratings increase, $D_{\mathrm{KL}}$ decreases for both models, but the trend is substantially sharper for the VLM. 
A linear fit on binned concreteness shows that the VLM’s alignment increases reliably with concreteness (slope$=-1.810$, $R^2=0.857$, $p<0.001$), whereas the LLM trend is weaker and not statistically reliable (slope$=-1.048$, $R^2=0.328$, $p>0.1$). 
This suggests that vision-text training does not merely shift ratings globally, but preferentially calibrates judgments for perceptually grounded words---precisely where vision provides an additional supervisory signal.

Because concreteness is an interpretable, human-normed semantic axis, improved human--model alignment makes model behavior easier to diagnose: it supports using concreteness as a principled factor for error analysis (e.g., when failures concentrate in abstract language) and as an interpretable control variable when comparing model families, aligning with calls for more rigorous, task-relevant interpretability evaluations \citep{doshi2017towards,Guidotti2019-GUIASO-3}. 
Mechanistically, concreteness alignment provides a concrete target for circuit-level analysis: one can localize where concreteness enters computation (layers/heads/MLP features) and test causal interventions, complementing transformer reverse-engineering frameworks \citep{elhage2021mathematical,geva-etal-2022-transformer}.

\paragraph{Model concreteness behaviour carries through scale.}
Across all analyses, the concreteness effects observed in the 8B model family qualitatively mirror in the 70B family: VLMs show larger gains on concrete QA, tighter within-concreteness clustering, more negative concreteness--entropy correlations, and stronger human-aligned rating trends. This consistency suggests that concreteness organization is not specific to small models, but a stable property that persists under scaling, with vision-text training providing an additive grounding signal rather than a scale-specific artifact.
\section{Conclusion}
We presented a controlled test of whether visual supervision during training induces more human-like concreteness sensitivity in foundation models. Holding the language backbone family and scaling regime as constant as possible and using text-only evaluation prompts, we compared matched Llama 3.1 LLMs against their Llama Vision 3.2 counterparts, treating the LLM--VLM contrast as an ablation on access to perceptual grounding rather than access to images at inference. Across diverse QA benchmarks, VLMs achieved higher accuracy overall and, crucially, showed larger improvements on questions with higher concreteness, consistent with grounded cognition and dual-coding accounts in which perceptual experience disproportionately supports concrete semantics \citep{paivio1990mental,barsalou2008grounded,harnad1990symbol}. Internal analyses converged on a coherent mechanistic picture: VLM token representations formed tighter within-concreteness clusters in low-dimensional projections, suggesting more stable type-level semantics for highly concrete words; and attention-entropy diagnostics indicated a sharper abstract--concrete separation in contextual reliance, aligning with psycholinguistic theories that abstract meaning draws more heavily on supportive context \citep{schwanenflugel-akin-luh-1992,schwanenflugel-shoben-1983}. Finally, elicited token-level concreteness ratings agreed more closely with human norms in VLMs, with a stronger improvement in human--model alignment as concreteness increased, indicating that vision--text training preferentially calibrates judgments precisely where perceptual supervision is informative.

Beyond the performance gap, our results position concreteness as a principled, interpretable axis for comparing model families and diagnosing grounding-related behavior.
This provides a useful bridge between cognitive constructs and modern interpretability practice: concreteness offers a measurable target for localizing where grounded semantic information enters computation.

\section*{Limitations}
\paragraph{Single model family.}
Our controlled comparison centers on one matched LLM/VLM family, improving internal validity but limiting generality. A direct next replication is to repeat the same pipeline on additional matched pairs, such as multilingual families Qwen/Qwen-VL to test whether concreteness effects and human-alignment patterns persist across languages and training recipes \citep{bai2023qwenvl,wang2024qwen2vl}.

\paragraph{Token frequency as a confound.}
Some apparent concreteness effects may be partly explained by lexical frequency/contextual diversity rather than grounding, but we cannot access true pretraining token counts. To reduce this confound, we can add frequency proxies (tokenizer-matched counts from large open corpora, plus average surprisal on held-out text) as covariates, or frequency-match concrete vs.\ abstract item sets before running the main regressions.

\paragraph{Developmental trajectory.}
We only analyze final checkpoints, so we cannot determine \emph{when} concreteness sensitivity and human-alignment emerge or how this depends on scale. An immediate follow-up is a developmental-style study over intermediate checkpoints (or staged training) to track the concreteness--accuracy slope and internal separability over training time for each model scale.



\bibliography{custom}

\appendix

\section{Model and Compute Details}
\label{app:model_details}

\paragraph{Model Architecture}
We evaluate models from the Llama 3.1 (text-only) and Llama 3.2 (vision-language) families. The vision-language models (VLMs) utilize the corresponding text-only models as their language backbone, augmented with a vision tower and cross-attention adapters. The architectural details for the text backbones are provided in Table~\ref{tab:model_specs}.

For the vision models (Llama 3.2 11B and 90B), the vision tower is a ViT-H/14 based encoder. The 11B VLM utilizes the 8B text backbone, while the 90B VLM utilizes the 70B text backbone. All models use Grouped-Query Attention (GQA) and are trained with a context window of 128k tokens.

\begin{table*}[h]
\centering
\small
{%
\begin{tabular}{lccccc}
\toprule
\textbf{Model Family} & \textbf{Size} & \textbf{Layers} & \textbf{Hidden Dim} & \textbf{Attn Heads} & \textbf{KV Heads} \\
\midrule
Llama 3.1 (Text) & 8B & 32 & 4096 & 32 & 8 \\
Llama 3.2 (Vision) & 11B & 32 & 4096 & 32 & 8 \\
\midrule
Llama 3.1 (Text) & 70B & 80 & 8192 & 64 & 8 \\
Llama 3.2 (Vision) & 90B & 80 & 8192 & 64 & 8 \\
\bottomrule
\end{tabular}%
}
\caption{Architectural specifications for the language backbones used in this study. The VLM variants inherit these text specifications and add vision encoder parameters.}
\label{tab:model_specs}
\end{table*}

\paragraph{Compute Resources}
All inference and evaluation experiments were conducted on a cluster of 8 \texttt{NVIDIA A40} GPUs. Models were loaded in \texttt{bfloat16} precision to match the training dtype. The total compute budget for the evaluation of all benchmarks and concreteness scoring was approximately \texttt{2 Days} of GPU hours.

\section{Dataset Details}
\label{app:datasets}

We utilize seven standard QA benchmarks covering distinct reasoning domains. Statistics for the evaluation splits used are detailed in Table~\ref{tab:dataset_stats}. For datasets where the official test set is hidden (BoolQ, WinoGrande, CommonsenseQA, PIQA), we report results on the canonical validation/development split. 
For ARC, we use the test split.
All datasets and model checkpoints are obtained from the public Hugging Face Hub (via the \texttt{datasets} and \texttt{transformers} libraries).
The benchmarks consist of generic multiple-choice questions and do not require any user-provided inputs.
We do not collect, store, or process personally identifying information. 
Accordingly, our experiments pose minimal risk of identity disclosure, and we report only aggregate accuracy metrics without releasing any per-example outputs that could contain sensitive content.


\begin{table*}[t]
\centering
\small
\begin{tabular}{llcc}
\toprule
\textbf{Dataset} & \textbf{Domain} & \textbf{\# Questions} & \textbf{Avg. Length} \\
\midrule
ARC-Easy        & Grade-school Science              & 2,376 & $\sim$39 words \\
ARC-Challenge   & Grade-school Science (Hard)        & 1,172 & $\sim$47 words \\
BoolQ           & Reading Comprehension (Yes/No)     & 3,270 & $\sim$120 words$^{\dagger}$ \\
WinoGrande      & Commonsense (Coreference)          & 1,267 & $\sim$32 words \\
CommonsenseQA   & General Commonsense                & 1,221 & $\sim$26 words \\
PIQA            & Physical Interaction               & 1,000 & $\sim$48 words \\
SIQA            & Social Commonsense                 & 1,954 & $\sim$36 words \\
\bottomrule
\end{tabular}
\caption{Summary of evaluation datasets. $^{\dagger}$Includes passage length.}
\label{tab:dataset_stats}
\end{table*}

\section{Prompts}
\label{app:prompts}

\paragraph{QA Evaluation (Non-Instruct Models)}
For the base (non-instruct) models, we utilized zero-shot prompting templates tailored to the task format. For multiple-choice datasets (ARC, CommonsenseQA, PIQA, WinoGrande), we used a standard completion format that lists options and prompts for an immediate answer:

\begin{quote}
\ttfamily
You are a helpful assistant. Answer the question immediately with just the option letter (A, B, C, D) or number. \\
Question: \{question\}\\
Options:\\
A. \{choice\_a\}\\
B. \{choice\_b\}\\
C. \{choice\_c\}\\
D. \{choice\_d\}\\
Answer:
\end{quote}

For the reading comprehension dataset (BoolQ), we employed a template that conditions the binary response on the provided passage:

\begin{quote}
\ttfamily
Read the following passage and answer the question with Yes or No.

Passage: \{passage\}\\
Questions: \{question\}\\
Answer:
\end{quote}

\paragraph{Concreteness Ratings}

To elicit concreteness ratings from the the large models (Llama 3.1 70B and Llama 3.2 90B Vision), we used the following prompt to ensure the output is aligned with the 1-7 MRC scale.

\begin{quote}
\ttfamily
You are a psycholinguistics expert. Your task is to rate the 'concreteness' of every content word in the following text on a scale from 1.0 (very abstract) to 7.0 (very concrete/tangible).\\

Defintion:\\
- **Concrete words** refer to things you can perceive directly with your senses (touch, see, hear, smell). Examples: 'apple', 'chair', 'scream'.\\
- **Abstract words** refer to concepts, ideas, or emotions that cannot be directly perceived. Examples: 'freedom', 'justice', 'infinity'.\\

Input Text: "\{text\}"\\

Return your analysis strictly as a JSON list of objects, where each object has 'word' and 'score'. Ignore stop words (the, a, is, etc.).\\
Example format:\\
{} [ \\
  \{"word": "apple", "score": 6.2\}, \\
  \{"word": "freedom", "score": 2.77\} \\
] \\
JSON Output:
\end{quote}

\section{Detailed Results}
\label{app:results}

Table~\ref{tab:per_dataset_acc} presents the raw accuracy scores for each model across all five datasets.

\begin{table*}[h]
\centering
\small
\setlength{\tabcolsep}{3pt} 
{%
\begin{tabular}{l cc | cc}
\toprule
& \multicolumn{2}{c|}{\textbf{Small Category}} & \multicolumn{2}{c}{\textbf{Large Category}} \\
\textbf{Dataset} & \textbf{Llama 3.1 8B} & \textbf{Llama 3.2 11B} & \textbf{Llama 3.1 70B} & \textbf{Llama 3.2 90B} \\
\midrule
ARC-Easy &86.45\% &\textbf{89.23\%} & 93.10\% & \textbf{93.90\%} \\
ARC-Challenge &65.70\% &\textbf{78.41\%} &89.08\% & \textbf{91.64\%} \\
BoolQ &76.18\% &\textbf{82.02\%} &85.02\% & \textbf{90.31\%} \\
WinoGrande & 52.41\%&\textbf{63.30\%} &69.93\% & \textbf{78.85\%} \\
CommonsenseQA &61.51\% &\textbf{71.58\%}&76.17\% & \textbf{77.15\%} \\
PIQA &35.30\% & \textbf{71.90\%} & 54.00\%& \textbf{74.50\%} \\
SIQA &64.43\% &\textbf{71.19\%} &65.10\% & 7\textbf{8.71\%} \\
\midrule
\textbf{Average} & & & & \\
\bottomrule
\end{tabular}%
}
\caption{Per-dataset accuracy (\%) for all evaluated models. The VLM variants generally perform comparable to or better than their text-only counterparts, despite receiving no visual input during this evaluation.}
\label{tab:per_dataset_acc}
\end{table*}

\section{Attention Entropy Analysis}
\label{sec:addi-entropy}
In the following tables, we present the raw correlation values from the Attention Entropy analysis for both model categories
Tables~\ref{tab:layer_stats_large_1}--\ref{tab:layer_stats_small} contain the Pearson correlation ($R$) and Significance ($p$) with significance levels: $^{**}p<0.05$, $^{***}p<0.01$.
\begin{table}[h]
\centering
\small
\setlength{\tabcolsep}{3pt} 
\caption{Large Scale Models (Part 1, Layers 0--31)}
\label{tab:layer_stats_large_1}
\begin{tabular}{l cc | cc}
\toprule
& \multicolumn{2}{c|}{\textbf{Llama 3.1 70B}} & \multicolumn{2}{c}{\textbf{Llama 3.2 90B}} \\
\textbf{Layer} & \textbf{$R$} & \textbf{$p$} & \textbf{$R$} & \textbf{$p$} \\
\midrule
0 & -0.20 & *** & -0.20 & *** \\
1 & -0.26 & *** & -0.22 & *** \\
2 & -0.26 & *** & -0.27 & *** \\
3 & -0.25 & *** & -0.30 & *** \\
4 & -0.15 & *** & -0.14 & *** \\
5 & -0.19 & *** & -0.26 & *** \\
6 & -0.24 & *** & -0.26 & *** \\
7 & -0.12 & *** & -0.15 & *** \\
8 & -0.19 & *** & -0.26 & *** \\
9 & -0.12 & *** & -0.19 & *** \\
10 & 0.04 & *** & -0.02 & 0.09 \\
11 & 0.00 & 0.82 & -0.10 & *** \\
12 & 0.06 & *** & 0.03 & *** \\
13 & -0.08 & *** & -0.15 & *** \\
14 & -0.03 & *** & -0.08 & *** \\
15 & -0.04 & *** & -0.11 & *** \\
16 & 0.04 & *** & -0.05 & *** \\
17 & -0.08 & *** & -0.11 & *** \\
18 & -0.08 & *** & -0.13 & *** \\
19 & -0.22 & *** & -0.27 & *** \\
20 & -0.17 & *** & -0.23 & *** \\
21 & -0.23 & *** & -0.30 & *** \\
22 & -0.15 & *** & -0.15 & *** \\
23 & -0.20 & *** & -0.23 & *** \\
24 & -0.24 & *** & -0.27 & *** \\
25 & -0.25 & *** & -0.24 & *** \\
26 & -0.27 & *** & -0.34 & *** \\
27 & -0.08 & *** & -0.18 & *** \\
28 & -0.06 & *** & -0.11 & *** \\
29 & -0.01 & 0.21 & -0.10 & *** \\
30 & 0.00 & 0.74 & -0.15 & *** \\
31 & -0.01 & 0.58 & -0.10 & *** \\

\bottomrule
\end{tabular}
\end{table}

\begin{table}[h]
\centering
\small
\setlength{\tabcolsep}{3pt}
\caption{Large Scale Models (Part 2, Layers 32--79)}
\label{tab:layer_stats_large_2}
\begin{tabular}{l cc | cc}
\toprule
& \multicolumn{2}{c|}{\textbf{Llama 3.1 70B}} & \multicolumn{2}{c}{\textbf{Llama 3.2 90B}} \\
\textbf{Layer} & \textbf{$R$} & \textbf{$p$} & \textbf{$R$} & \textbf{$p$} \\
\midrule

32 & 0.05 & *** & -0.08 & *** \\
33 & 0.05 & *** & -0.14 & *** \\
34 & 0.08 & *** & -0.08 & *** \\
35 & 0.00 & 0.78 & -0.13 & *** \\
36 & 0.00 & 0.84 & -0.16 & *** \\
37 & 0.13 & *** & 0.02 & ** \\
38 & 0.15 & *** & 0.05 & *** \\
39 & 0.11 & *** & 0.03 & ** \\
40 & 0.20 & *** & 0.07 & *** \\
41 & -0.05 & *** & -0.22 & *** \\
42 & -0.01 & 0.27 & -0.14 & *** \\
43 & 0.08 & *** & -0.09 & *** \\
44 & 0.12 & *** & -0.02 & 0.06 \\
45 & 0.19 & *** & 0.08 & *** \\
46 & -0.02 & 0.09 & -0.07 & *** \\
47 & 0.11 & *** & 0.00 & 0.88 \\
48 & 0.08 & *** & -0.03 & *** \\
49 & 0.10 & *** & -0.04 & *** \\
50 & -0.00 & 0.87 & -0.09 & *** \\
51 & 0.08 & *** & -0.03 & *** \\
52 & 0.14 & *** & 0.02 & 0.15 \\
53 & 0.06 & *** & -0.02 & 0.06 \\
54 & 0.08 & *** & -0.06 & *** \\
55 & 0.02 & ** & -0.15 & *** \\
56 & 0.15 & *** & 0.06 & *** \\
57 & 0.07 & *** & -0.03 & *** \\
58 & -0.06 & *** & -0.11 & *** \\
59 & -0.12 & *** & -0.19 & *** \\
60 & 0.14 & *** & 0.04 & *** \\
61 & -0.15 & *** & -0.15 & *** \\
62 & -0.16 & *** & -0.16 & *** \\
63 & -0.02 & 0.09 & -0.11 & *** \\
64 & 0.02 & 0.11 & -0.02 & 0.16 \\
65 & -0.14 & *** & -0.15 & *** \\
66 & -0.04 & *** & -0.08 & *** \\
67 & 0.09 & *** & -0.00 & 0.78 \\
68 & 0.09 & *** & 0.02 & ** \\
69 & -0.00 & 0.83 & -0.06 & *** \\
70 & 0.10 & *** & 0.03 & *** \\
71 & 0.11 & *** & 0.02 & ** \\
72 & 0.14 & *** & 0.03 & *** \\
73 & 0.07 & *** & -0.04 & *** \\
74 & 0.11 & *** & 0.06 & *** \\
75 & 0.10 & *** & 0.04 & *** \\
76 & 0.03 & *** & -0.03 & *** \\
77 & 0.06 & *** & 0.02 & ** \\
78 & 0.00 & 0.82 & -0.01 & 0.31 \\
79 & -0.07 & *** & -0.08 & *** \\
\bottomrule
\end{tabular}
\end{table}

\begin{table}[h]
\centering
\small
\setlength{\tabcolsep}{3pt} 
\caption{Small Scale Models (Layers 0--31)}
\label{tab:layer_stats_small}
\begin{tabular}{l cc | cc}
\toprule
& \multicolumn{2}{c|}{\textbf{Llama 3.1 8B}} & \multicolumn{2}{c}{\textbf{Llama 3.2 11B}} \\
\textbf{Layer} & \textbf{$R$} & \textbf{$p$} & \textbf{$R$} & \textbf{$p$} \\
\midrule
0 & -0.20 & *** & -0.20 & *** \\
1 & -0.32 & *** & -0.28 & *** \\
2 & -0.24 & *** & -0.25 & *** \\
3 & -0.27 & *** & -0.31 & *** \\
4 & -0.27 & *** & -0.31 & *** \\
5 & -0.23 & *** & -0.28 & *** \\
6 & -0.29 & *** & -0.31 & *** \\
7 & -0.21 & *** & -0.25 & *** \\
8 & -0.17 & *** & -0.19 & *** \\
9 & -0.17 & *** & -0.19 & *** \\
10 & -0.13 & *** & -0.17 & *** \\
11 & -0.10 & *** & -0.13 & *** \\
12 & -0.18 & *** & -0.22 & *** \\
13 & -0.18 & *** & -0.19 & *** \\
14 & -0.15 & *** & -0.22 & *** \\
15 & -0.05 & *** & -0.12 & *** \\
16 & 0.03 & *** & -0.01 & 0.19 \\
17 & -0.01 & 0.27 & -0.11 & *** \\
18 & -0.10 & *** & -0.16 & *** \\
19 & 0.01 & 0.32 & -0.07 & *** \\
20 & -0.01 & 0.24 & -0.09 & *** \\
21 & -0.03 & *** & -0.11 & *** \\
22 & -0.01 & 0.38 & -0.08 & *** \\
23 & -0.09 & *** & -0.17 & *** \\
24 & 0.01 & 0.28 & -0.02 & ** \\
25 & -0.04 & *** & -0.09 & *** \\
26 & 0.01 & 0.22 & -0.02 & ** \\
27 & -0.01 & 0.25 & -0.06 & *** \\
28 & -0.07 & *** & -0.10 & *** \\
29 & -0.13 & *** & -0.12 & *** \\
30 & -0.07 & *** & -0.09 & *** \\
31 & -0.08 & *** & -0.04 & *** \\
\bottomrule
\end{tabular}
\end{table}


\end{document}